\documentclass{article}
\usepackage{arxiv}

\usepackage[utf8]{inputenc} 
\usepackage[T1]{fontenc}    
\usepackage{hyperref}       
\usepackage{url}            
\usepackage{booktabs}       
\usepackage{amsfonts}       
\usepackage{nicefrac}       
\usepackage{microtype}      
\usepackage{lipsum}
\usepackage{graphicx}
\usepackage{float}
\usepackage{algorithm}
\usepackage{algpseudocode}
\usepackage{amsmath}
\usepackage{tikz}
\usepackage{caption,subcaption}
\usepackage{makecell}
\newcommand{\norm}[1]{\left\lVert#1\right\rVert}

\graphicspath{ {./images/} }

\title{Actor-Critic Algorithm for High-dimensional\\ Partial Differential Equations}

\author{
 Xiaohan Zhang \\
  \texttt{xiaohan.zhang@salesforce.com} \\
}

\begin{document}
\maketitle
\begin{abstract}

We develop a deep learning model to effectively solve high-dimensional nonlinear parabolic partial differential equations (PDE). We follow Feynman-Kac formula to reformulate PDE into the equivalent stochastic control problem governed by a Backward Stochastic Differential Equation (BSDE) system. The Markovian property of the BSDE is utilized in designing our neural network architecture, which is inspired by the Actor-Critic algorithm usually applied for deep Reinforcement Learning. Compared to the State-of-the-Art model, we make several improvements including 1) largely reduced trainable parameters, 2) faster convergence rate and 3) fewer hyperparameters to tune. We demonstrate those improvements by solving a few well-known classes of PDEs such as Hamilton-Jacobian-Bellman equation, Allen-Cahn equation and Black-Scholes equation with dimensions on the order of 100.

\end{abstract}


\section{Introduction}
High Dimensional partial differential equations (PDEs) are encountered in many branches of modern sciences such as the Schr\"{o}dinger equation for quantum many-body problem, the nonlinear Black-Scholes equation for pricing financial derivatives, and the Hamilton-Jacobi-Bellman equation for multi-agent game theories, to name a few. The ``curse of dimensionality`` is a major roadblock for generalizing classical low dimensional algorithms to the high dimension settings (say hundreds of dimensions). Namely, as the dimensionality grows, the complexity of the algorithms grows exponentially. For this reason, there exist only a very limited number of cases where practical high dimensional algorithms have been developed. We refer to \cite{weinan2017deep} for an exhaustive review of those cases. 

 Recently, deep learning based PDE solvers become popular, which is led by the practical success of deep neural network in dealing with other high dimensional problems such as computer vision and natural language processing. Han and E \cite{weinan2017deep,han2018solving} have developed a leading-edge  ``deep Backward Stochastic Differential Equation'' model (DBSDE). The key ideas of DBSDE involve 1) reformulating PDEs into the equivalent stochastic control problem governed by a backward stochastic differential equation(BSDE) system; 2) making an analogy between BSDE and  model-based reinforcement learning, where the gradients of the solution playing the role of a policy function; 3) parametrizing the policy function by a time-sequence of multilayer feedforward neural networks. The accuracy and stability of DBSDE have been demonstrated in \cite{weinan2017deep} and several follow-up studies \cite{han2018solving,pham2019neural,liang2019deep}. 
 
 In spite of being the State-of-the-Art solver for high dimensional PDE, DBSDE has a few issues that limit it from practicability. In this work, we call attention to those issues and introduce a new model framework to address them. The design of our model is inspired by the Actor-Critic algorithm which is usually seen in Reinforcement Learning problems. The following list compiles some key improvements realized by our model compared to the State-of-the-Art:
 
\begin{enumerate}
    \item {\emph{largely reduced trainable parameters from $\mathcal{O}(Nd^2)$ to $\mathcal{O}(d^2)$}:} here $N$ is the number of time steps that discretizes the temporal dimension, and $d$ is the spatial dimension of the PDEs. Namely, we get rid of the constraint that the network complexity need to scale linearly with the time steps. The immediate benefits of having a light-weight neural network are faster iteration speed and less memory consumption. 
    \item {\emph{faster convergence rate:}} In all the numerical experiments we studied, the convergence rate of our model is generally an order faster than DBSDE while giving the same (if not higher) level of solution accuracy. Combining with the previous improvement on reducing parameters, our model guarantees a much shorter run-time in solving the PDEs compared to DBSDE. For example, Quadratic Gradients equation is solved $\sim 18$ times faster and the Allen Cahn equation is $\sim 27$ times faster.  
    \item {\emph{less hyperparameters to tune:}} DBSDE expects two hyperparameters that define a range from which the initial solution can be sampled. our model dispenses such need and consequently saves the time in hyperparameter tuning. This improvement is automatically enabled by the Actor-Critic setup of the neural networks.
\end{enumerate}

We close this section by mentioning some details about the numerical implementations. We implement both our model and the DBSDE model under the same deep reinforcement learning framework  \footnote{PyTorch version of spinningup (https://spinningup.openai.com) to implement both algorithms.} to maximally eliminate confounding factors when comparing the algorithms. We have also reproduced all the results presented in this work using the original DBSDE code which was written in TensorFlow. 

\section{Method}\label{sec:method}
We consider the following nonlinear parabolic PDE:
\begin{equation}\label{eqn:pde}
    \frac{\partial u}{\partial t} \left(t,x\right) + \frac{1}{2}{\rm Tr}\left( \sigma \sigma^T\left(t,x\right)(\rm{Hess}_x u)(t,x)\right) + \nabla u \left(t,x\right)\cdot\mu\left(t,x\right) + f\left(t,x,u(t,x),\sigma^T(t,x)\nabla u(t,x)\right)=0
\end{equation}
with terminal condition $u(T,x)=g(x)$. Here $t\in [0,T]$ and $x\in \mathbb{R}^d$ are the time and space variable respectively. $\mu (t,x) \in \mathbb{R}^d$ and $\sigma(t,x) \in \mathbb{R}^{d\times d}$ are known vector-valued functions. $\sigma^T$ is the transpose of $\sigma$.  $\nabla u$ and ${\rm{Hess}}_x u$ represents the gradient and the Hessian of function $u$ w.r.t $x$. Tr denotes the trace of a $d\times d$ matrix. $f$ is a known scalar-valued nonlinear function. The goal is to find the solution $u(0,\xi)$ for some point $\xi\in \mathbb{R}^d$ at $t=0$. It is worthy of pointing out that many well-known PDEs are all particular forms of the equation \eqref{eqn:pde}: Schr\"{o}dinger equation, Hamilton-Jacobian-Bellman equation, Allen-Cahn equation, Black-Scholes equation and Burgers Type Equation, to name a few. 

We follow Feynan-Kac formula to draw the connection between the PDE  (\ref{eqn:pde}) and its equivalent stochastic control problem. To fix some terminology, let $(\Omega,\mathcal{F}, \mathbb{P})$ be a probability space, $W:[0,T]\times \Omega \rightarrow \mathbb{R}^d$ be a d-dimensional standard Brownian motion, $\mathcal{F}_{t\in[0,T]}$ be the normal filtration generated by ${W}_{t\in[0,T]}$. Consider the $\mathcal{F}_{t\in[0,T]}$-adapted solution process $(\{X_t, Y_t, Z_t\}_{t\in[0,T]})$ of the following Backward Stochastic Differential Equation (BSDE) system:
\begin{equation}
\label{eqn:bsde}
\left\{
\begin{aligned}
&X_t =  \xi + \int^t_0 \mu(s, X_s)\,ds + \int^t_0 \sigma(s,X_s)\,dW_s\\
&Y_t =  g(X_T) + \int^T_t f(s, X_s,Y_s,Z_s)\,ds -\int^T_t (Z_s)^T\,dW_s\\
\end{aligned}
\right.
\end{equation}
where $\{X_t \}_{t\in\left[0,T\right]}$ is a $d-$dimensional stochastic process. The solution process is proved to exist with up-to-indistinguishability uniqueness \cite{pardoux1992backward, el1997backward}) under suitable regularity assumptions on the coefficient functions $\mu$, $\sigma$, and $f$. As a result of Feynman-Kac formula (cf., e.g., \cite{pardoux1992backward, pardoux1999forward, el1997backward}), the solution process $(\{X_t, Y_t, Z_t\}_{t\in[0,T]})$  is related to $u(t,x)$ in the sense that for all $t\in[0,T]$ it holds $\mathbb{P}$-a.s. that
\begin{equation}\label{eqn:identity}
\begin{aligned}
    &Y_t = u(t,X_t), \quad\text{and}\quad 
    Z_t = \sigma^T(t,X_t)\nabla u(t,X_t)
\end{aligned}
\end{equation}
Plugging (\ref{eqn:identity}) into (\ref{eqn:bsde}) and rewriting (\ref{eqn:bsde}) forwardly, we obtain the following stochastic system:
\begin{equation}\label{eqn:uxz}
\begin{aligned}
    &X_t = \xi + \int^t_0\mu(s,X_s)\,ds + \int^t_0\sigma(s,X_s)\,dW_s, \quad Z_t = \sigma^T(t,X_t)\nabla(t,X_t) \\
    &u(t,X_t) = u(0,X_0) - \int^t_0 f\left(s, X_s, u(s,X_s), Z_s\right)\, ds + \int^t_0 Z^T_s\,dW_s
\end{aligned}
\end{equation}
 So far, we have turned the problem of solving the PDE  \eqref{eqn:pde} into solving a backward stochastic differential equation (BSDE) system. To solve the system, we first discretize all the stochastic processes in time with a simple Euler scheme. Namely, with a partition of the time interval $[0,T]: 0 < t_0 <t_1 <... <t_N$, we have the discretized form:
\begin{subequations}\label{eqn:uxzn}
\begin{align}
& X_{t_{n+1}} = X_{t_n} + \mu(t_n, X_{t_n})(t_{n+1}-t_n)+\sigma(t_n, X_{t_n})(W_{t_{n+1}}-W_{t_n}) \label{eqn:uxzn:a}\\
& u(t_{n+1},X_{t_{n+1}}) = u(t_n, X_{t_n}) - f\left(t_n, X_{t_n}, u(t_n, X_{t_n}), Z_{t_n} \right) + Z_{t_n}^T(W_{t_{n+1}}-W_{t_n}),\label{eqn:uxzn:b}
\end{align}
\end{subequations}
where $Z_{t_n}=\left[\nabla u(t_n, X_{t_n}) \right]^T\sigma(t_n, X_{t_n})$. From the numerical point of view, \eqref{eqn:uxzn:a} defines a controlled stochastic dynamics that can be efficiently sampled by simulating Brownian processes $W_{t_n}$, with $\mu$ and $\sigma$ given. Note that $N$ is a hyperparameter which needs to be tuned for different equations. The sensitivity study of $N$ is yet available in the literature.

A key feature that differentiates our model from others is that we leverages the Markovian property of the System (\ref{eqn:uxzn}). Therefore, we only deploy {\emph{one}} multilayer feedforward network with batch-normalization, $\theta_a$, to approximate $Z_{t_n}:n=1,2,\ldots N$, whereas a sequence of $N$ networks are adopted by DBSDE and other models alike. To some extent, $\theta_a$ behaves similarly to the policy network in model-based reinforcement learning. In addition, we parametrize $u(X_{t=0})$ with a second multilayer feedforward network, $\theta_v$, similar to the critic network in reinforcement learning. A combination of such two networks within one framework is normally referred to as Actor-Critic algorithm which has been extensively applied to Markovian Decision Process (or reinforcement learning in general). 

\begin{figure}[H] 
    \centering
    \includegraphics[width=.45\columnwidth]{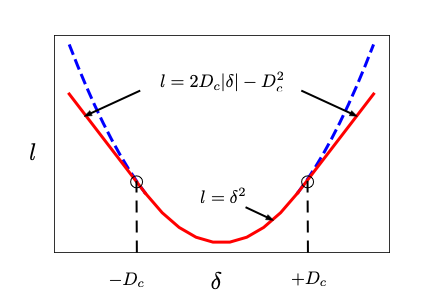}
    \caption{Sketch of minimization objective as a function of $\delta=\left|g(X_T)-u(\{X_{t_n}\}_{0\leq n\leq N},\{W_{t_n}\}_{0\leq n\leq N}) \right|$. The function is quadratic within $[-D_c,D_c]$ while linear beyond, to prevent the loss from exploding.}
    \label{fig:loss}
\end{figure}

To close the loop, we still need to define a loss function for training: 
\begin{equation}
    l(t=T;\theta_v, \theta_a) = \mathbb{E}\left[ \left|g(X_T)-u(\{X_{t_n}\}_{0\leq n\leq N},\{W_{t_n}\}_{0\leq n\leq N}) \right|^2\right]
\end{equation}
namely the loss function measures how close the predicted solution $u(T,x)$ matches the terminal boundary condition. In practice, to prevent the loss from blowing up, we clip the quadratic function by linearly extrapolating the function beyond a predefined domain $[-D_c,D_c]$, analogous to the trick used by Proximal Policy Optimization \cite{schulman2017proximal} which enforces a not-too-far policy update  \footnote{The difference is that PPO puts the constraint on the KL-divergence between consecutive updates instead of the least square measure.}. We use $D_c=50$ in all of our experiments. Fig. \ref{fig:loss} illustrates the linear-clipping trick.

\begin{figure}[H] 
    \centering
    \includegraphics[width=.618\columnwidth]{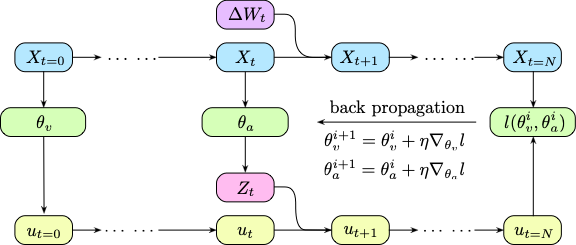}
    \caption{Forward and backward propagation of the $i_{\text{th}}$ iteration}
    \label{fig:algorithm}
\end{figure}

Given the temporal discretization above, the path $\{X_{t_n}\}_{0\leq n\leq N}$ can be easily sampled using (\ref{eqn:uxzn:a}), the dynamics of which are problem dependent due to those $\mu$ and $\sigma$ terms.  Fig. \ref{fig:algorithm} shows the relationship between the policy and critic networks, as well as other components that go into the system. The flow chart illustrates a forward pass and a backward pass in one iteration where $\theta_v$ and $\theta_a$ are updated.

The training process is summarized by Algorithm \ref{alg:ppo} below. Essentially, for each iteration, $X_t$, $W_t$ are sampled first. We use $\theta_v$ to generate a guess, $u(X_{t=0})$ which is then passed forward in time to get $u(X_{t=T})$. The loss is backpropagated to update $\theta_v$ and $\theta_a$ with either stochastic gradient descent or other  methods alike. The total number of training steps is preset but we also find that using an early-stop mechanism usually leads to shorter run time while producing the same level of accuracy. \footnote{We do not use early-stopping in the numerical experiments as we want to have a fair comparison with other models in terms of run-time and convergence rate.}

\begin{figure}[H]
\centering
\begin{minipage}{.9\linewidth}
\begin{algorithm}[H]
\caption{Training Process}
\label{alg:ppo}
	\begin{algorithmic}[1]
	\State {Input: initialize policy network $\theta_a$ and critic network $\theta_v$ with Xavier uniform initializer.}
		\For {$iteration=1,2,\ldots$}
				\State Collect set of trajectories of $X_{t=1\ldots N}$ and $W_{t=1\ldots N}$ by running \eqref{eqn:uxzn:a}
				\State Compute $u(X_{t=0})$ from critic network with $\theta_v$. 
				\For {$t=1,2,\ldots$,N}
				    \State Compute $u(X_{t+1})$ from $u(X_{t})$, policy network $\theta_a$ and function $f$.
				\EndFor
			\State Compute loss value from $u(X_{t=N})$ and $g(X_T)$
            \State Minimize the loss and update $\theta_v$ and $\theta_a$, typically via stochastic gradient descent.
		\EndFor
	\end{algorithmic} 
    \end{algorithm}
  \end{minipage}
\end{figure}

\section{Results}
\label{sec:results}
 We solve all the examples presented by \cite{weinan2017deep, han2018solving} using our model and compare the results with the State-of-the-Art DBSDE model. Table \ref{tab:eqn_detail} lists the choices of the examples. We refer the readers to Appendix \ref{appendix:pde} for more details about the values of the parameters and the boundary conditions etc. It is important to mention that we intentionally use the same hyperparameters and settings whenever possible, in order to pinpoint the algorithm advantage. 

\begin{table}[H]
 \caption{PDE examples studied in this paper and in  \cite{weinan2017deep,han2018solving}}
  \centering
  \begin{tabular}{llll}
    \toprule
    PDE Classes    & $\sigma(t,x)$ & $\mu(t,x)$ & $f\left(t,x,u(t,x),\sigma^T\nabla u(t,x)\right)$ \\
    \midrule
    Hamilton Jacobi Bellman& $\sqrt{2}$ & 0 & $-\norm{z}_{\mathbb{R}^{1\times d}}^{2}$  \\
    Burgers Type            & $\frac{d}{\sqrt{2}}$ &$0$ & $\left(y-\frac{2+d}{2d} \right)\left(\sum_{i=1}^d z_i\right)$ \\
    Reaction Diffusion     & $1$ & $0$ & 
    $\begin{aligned}
    \min\left\{1,\left[y-\kappa-
    \sin(\lambda\sum^d_{i=1}x_i) e^{\frac{\lambda^2d(t-T)}{2}}\right]^2\right\}
    \end{aligned}$  \\
    Quadratic Gradients    & $1$&$0$ & $
    \begin{aligned}
    \norm{z}^2_{\mathbb{R}^{1\times d}} - \norm{(\nabla_x\psi)}^2_{\mathbb{R}^d} -
    \psi_t - \frac{1}{2}(\Delta_x \psi)
    \end{aligned}$  \\
    Allen Cahn             & $\sqrt{2}$ & $0$ & $y-y^3$  \\
    Pricing Option         & $\Bar{\sigma}{\rm diag}(x_1, \dots, x_d)$& $\Bar{\mu}x$ & $ \begin{aligned}  -R^l y-\frac{\Bar{\mu}-R^l}{\Bar{\sigma}}\sum^d_{i=1}z_i +   R^b
        \norm{\frac{1}{\Bar{\sigma}}\sum^d_{i=1}z_i-y}_\infty
    \end{aligned}$  \\
    \bottomrule
  \end{tabular}
  \label{tab:eqn_detail}
\end{table}

We use relative error with respect to the ``exact'' solutions for each equation to validate our model accuracy. Namely, we can compute
\begin{equation}\label{eqn:relative_error}
 \text{\rm Relative Error} = \frac{u_{\text{exact}}(0,\xi)-u_{\text{predicted}}(0,\xi)}{u_{\text{exact}}(0,\xi)}   
\end{equation}
for both DBSDE and our Actor-Critic model. We refer to the Appendix for more details on obtaining the ``exact'' solutions for each equation.

For each of the equation in Table \ref{tab:eqn_detail}, we perform 5 independent training and monitor the solution $u(0,\xi)$, the loss function $l(\theta_v, \theta_a)$ and relative approximation error \eqref{eqn:relative_error} during training. When plotting the training histories, we use a shadow area to indicate the variance of the 5 runs while solid line indicate the mean. 

All experiments are performed in Python using PyTorch on a Google Colab server which has 4 cores of Intel Xeon micro processor of 2000 Megahertz (MHz) and one Tesla V100-SXM2 GPU.  We reiterate that DBSDE is implemented and run in the same environment such that the difference in the performance stems from the algorithm designs exclusively. 

In the rest of this section, we systematically compare 1) the run-time and accuracy 2) the number of trainable parameters, 3) the convergence rate and 4) the number of hyperparameters between our model and DBSDE. The goal is to provide numerical evidence to our model advantages that are enumerated in Section \ref{sec:method}. 

\subsection{Model effectiveness - shorter run time with same accuracy}\label{sec:results_accuracy_runtime}

 The key results about the model effectiveness are presented by table \ref{tab:main_results}. Basically, we can solve all the examples at least one order of magnitude faster than DBSDE. For example, the Quadratic Gradients equation is solved $\sim18$ times faster and Allen Cahn is $\sim27$ times faster.  Furthermore, in 5 out of 6 experiments, we achieve even higher accuracy than DBSDE. For completeness, we also include the relative error of DBSDE reported by \cite{weinan2017deep}.  We can achieve better run-time performance because our model allows a reduced number of network parameters and faster convergence rate, which will be discussed in Section \ref{sec:results_train_params}  and \ref{sec:results_convergence_rate} respectively.

It is important to mention that we use the same set of  hyperparameters such as learning rate scheduler, choice of activation functions, batch size, time steps etc for both models. Those parameters are only optimized for DBSDE to make the comparison straightforward.

\begin{table}[H]
 \caption{Run-time and Relative-error for all numerical examples}
  \centering
  \begin{tabular}{lrrrrr}
    \toprule
    \thead{PDE Examples}    & \thead{Run Time \\Actor-Critic} & \thead{Run Time \\(DBSDE)} & \thead{Relative Error\\Actor-Critic} & \thead{Relative Error \\(DBSDE)} & \thead{Relative Error \\ reported by \cite{weinan2017deep}}\\
    \midrule
    Hamilton Jacobi Bellman & $3\,\rm s$& $22\,\rm s$& 0.22\% & 0.53\% & 1.7\% \\
    Burgers Type            & $20\,\rm s$ & $122\,\rm s$ & 3.4\% &0.31\% & 0.35\% \\
    Reaction Diffusion      & $132\,\rm s$ & $801\,\rm s$ & 0.61\% & 0.69\%  & 0.60\%\\
    Quadratic Gradients     & $9\,\rm s$ & $166\,\rm s$ & $0.06\%$ & $0.08\%$ & $0.09\%$ \\
    Allen Cahn              & $5\,\rm s$ & $138\,\rm s$ & 0.25\% & 0.46\% & 0.30\% \\
    Pricing Option          & $7\,\rm s$ & $20\,\rm s$ & 0.37\% & 0.56\%  & 0.40\% \\
    \bottomrule
  \end{tabular}
  \label{tab:main_results}
\end{table}

\subsection{Reduced trainable parameters}\label{sec:results_train_params}

we use a fully-connected (FC) feedforward neural networks with batch normalization to represent $\theta_a$ and another FC for $\theta_v$. Each of the neural networks consists of 4 layers (1 input layer of $d$-dimensional, 2 hidden layers of both $d+10$-dimensional, and 1 output layer of $d$-dimensional). The number of hidden units in each hidden layer is equal to $d + 10$. We employ ReLU as our activation function. All the weights in the network are initialized with the standard Glorot initialization (also known as xavier uniform initialization) \cite{glorot2010understanding} without any pre-training.

The number of trainable parameters in our algorithm, $\rho_0$, can be calculated as:
\begin{equation}\label{eq:actor_critic_param}
    \rho_0 = 2\times \underbrace{\left((d + 10) + (d + 10)^2 + d(d + 10)\right) }_{\text{fully connected layers of $\theta_a$ and $\theta_v$}} + \underbrace{2(d+10)+ 2(d+10)+ 2d}_{\text{batch normalization layers of $\theta_a$ and $\theta_v$}}
\end{equation}
In comparison, the number of trainable parameters of DBSDE model is calculated as:
\begin{equation}\label{eq:DBSDE_param}
\begin{aligned}
\rho_1 &= \underbrace{1+d}_{u(0,\xi), \nabla u(0,\xi)} +  \underbrace{(N-1)(d(d+10)+(d+10)^2+d(d+10))}_{\text{fully connected layers}} \\
&+\underbrace{(N-1)(2(d+10)+2(d+10)+2d)}_{\text{batch normalization layers}}
\end{aligned}
\end{equation}
Comparing \eqref{eq:actor_critic_param} and \eqref{eq:DBSDE_param}, one  immediately notice that:
\begin{enumerate}
    \item DBSDE uses one parameter to approximate $u(0,\xi)$ and $d$ parameters for $\nabla(0,\xi)$. We do not have those two sets of parameters.
    \item The network proposed by DBSDE is a MLP stacked $N$ times where $N$ is the time steps that discretize the temporal dimension, which leads to $\rho_1\sim\mathcal{O}(Nd^2)$ complexity, whereas  $\rho_0\sim\mathcal{O}(d^2)$ in our model. Recall that $N$ is a hyperparameter that needs to be tuned case by case. Therefore, having the network complexity controlled by $N$ poses numerical challenges when $N$ is large. Our model has no such constraint.
\end{enumerate}

Let P represent the number of trainable parameters and RpI represent the run time per iteration.  Table \ref{tab:param_count} shows the ratio between our model and DBSDE for P and RpI in solving all six equations.

\begin{table}[H]
 \caption{Number of parameters (${\rm P}$) and run-time per iteration (${\rm RpI}$).}
  \centering
  \begin{tabular}{lllrr}
    \toprule
    PDE Examples    & Dimension & Time Step &${\rm P/P_{\text{DBSDE}}}$& ${\rm RpI/RpI_{\text{DBSDE}}}$ \\
    \midrule
    Burgers Type            & $d=50$ &$N=30$ & $6.5\%$ & $44.4\%$ \\
    Reaction Diffusion     & $d=100$&$N=30$ & $6.7\%$ & $51.1\%$ \\
    Quadratic Gradients    & $d=100$&$N=30$ & $6.7\%$ & $48.7\%$  \\
    Hamilton Jacobi Bellman& $d=100$&$N=3$ & $96.8\%$ & $58.8\%$  \\
    Allen Cahn             & $d=100$& $N=20$ & $10.2\%$ & $41.6\%$  \\
    Pricing Option         & $d=100$& $N=20$ & $10.2\%$ & $47.2\%$  \\
    \bottomrule
  \end{tabular}
  \label{tab:param_count}
\end{table}

We observe a significant reduction in the number of trainable parameters comparing to DBSDE, especially for the cases where $N$ is large. The run-time per iteration is also shorter in our model, although the speedup factor is not necessarily proportional to the reduction in parameter sizes. 

\subsection{Faster Convergence Rate}\label{sec:results_convergence_rate}
\begin{figure}[H]
    \centering 
\begin{subfigure}{0.25\textwidth}
  \includegraphics[width=\linewidth]{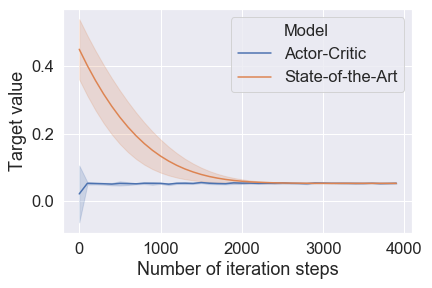}
  \caption{Allen Cahn}
  \label{fig:1}
\end{subfigure}\hfil 
\begin{subfigure}{0.25\textwidth}
  \includegraphics[width=\linewidth]{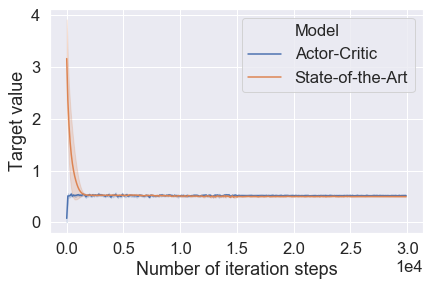}
  \caption{Burgers Type}
  \label{fig:2}
\end{subfigure}\hfil 
\begin{subfigure}{0.25\textwidth}
  \includegraphics[width=\linewidth]{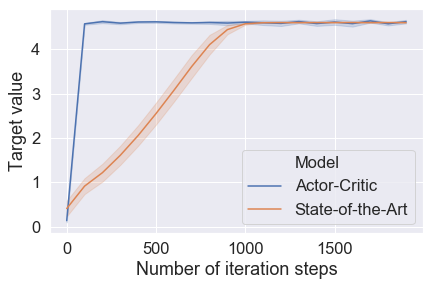}
  \caption{Hamilton Jacobian Bellman}
  \label{fig:3}
\end{subfigure}

\medskip
\begin{subfigure}{0.25\textwidth}
  \includegraphics[width=\linewidth]{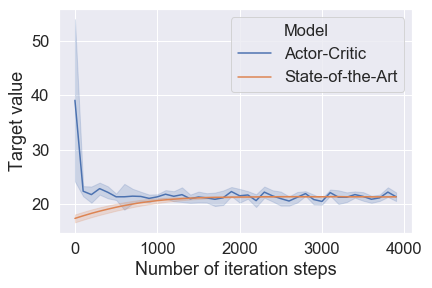}
  \caption{Pricing Option}
  \label{fig:4}
\end{subfigure}\hfil 
\begin{subfigure}{0.25\textwidth}
  \includegraphics[width=\linewidth]{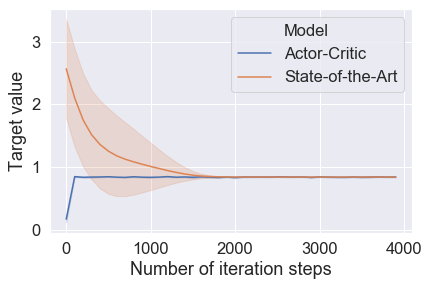}
  \caption{Quadratic Gradients}
  \label{fig:5}
\end{subfigure}\hfil 
\begin{subfigure}{0.25\textwidth}
  \includegraphics[width=\linewidth]{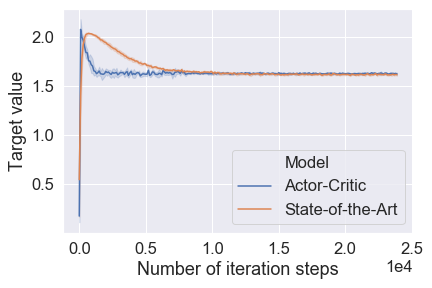}
  \caption{Reaction Diffusion}
  \label{fig:6}
\end{subfigure}
\caption{Evolution of the target solution $u(0,\xi)$ during training}
\label{fig:target_solution}
\end{figure}

For each of the 5 independent training, we randomly initialize the network parameters, and choose different random seeds for generating the Brownian motion sample paths. A training is considered to finish when an ``equilibrium'' state is reached, i.e., either the fluctuations of $u(0,\xi)$ or the loss remain sufficiently small (with a threshold $\epsilon$) for a long period (with a threshold $P$). We refer to the appendix for the specific choice of $\epsilon$ and $P$ in each case. The convergence rate is defined as the number of iterations that the training takes to reach the equilibrium. In Figure \ref{fig:target_solution}, we compare the convergence rate from DBSDE and our model when solving the equations in table \ref{tab:eqn_detail}. In addition to the convergence rate, we also present the training history of the loss function, relative approximation error in Appendix \ref{appendix:training}.

Compared to DBSDE, we notice that our model needs much fewer iterations to converge. Intuitively, this could be attributed to the fact that our neural network is shallower than DBSDE by design, which usually requires fewer samples to train. Recall that our run-time for each iteration is also shorter than DBSDE, as shown in Table \ref{tab:param_count}. Both effects combined explain why our training speed is much faster than DBSDE as previously shown in Table \ref{tab:main_results}.  

\subsection{Two fewer hyperparameters}
DBSDE treats the solution $u(0,\xi)$ as a trainable parameter which needs to be initialized properly. To solve all the equations with the reported accuracy in table \ref{tab:main_results}, DBSDE model requires tuning two hyperparameters, say lo and hi, from case to case. Here, lo and hi are two scalars that define the range $[\text{lo}, \text{hi}]$ from which the initial guess of $u(0,\xi)$ is sampled. The process of tuning lo and hi can be overwhelming. To reproduce the results presented by \cite{weinan2017deep,han2018solving}, we set up a grid search to obtain the best combination of lo and hi. For example, we use [lo=0.3, hi=0.6] for Allen-Cahn equation, [lo=15, hi=18] for Pricing Option equation, [lo=2, hi=4] for Burgers Type equation, and [lo=0, hi=1] for Reaction Diffusion equation. A careless choice of lo and hi sometimes leads to poorer accuracy or even a divergent loss. 

 In contrast to DBSDE, we use the critic network to approximate $u(0,\xi)$. We use the same critic network structure and same initialization procedure (xavier-uniform) in solving all the equations of table \ref{tab:eqn_detail}. To some extent, the experiments suggest that the critic network, if properly designed and initialized, is more appropriate in regularizing $u(0,\xi)$ than using one parameter.  

\section{Conclusions}
We have developed a deep learning model to effectively solve high dimensional parabolic Partial Differential Equations. Our algorithm design is inspired by the Actor-Critic framework commonly used by deep reinforcement learning models. Through  numerical experiments where we compare our results with the State-of-the-Art, our model is demonstrated to persist several advantages including reduced trainable parameters, faster converge rate and fewer hyperparameters to tune. As a result, our model solves the PDEs at least one order of magnitude faster preserving the same accuracy level. 
\bibliographystyle{unsrt}  
\bibliography{nnpde}  

\appendix
\section{Feyman-Kac Formula}\label{appendix:feyman-kac}
Let $Y:[0,T]\times \Omega \rightarrow \mathbb{R}$ and $Z:[0,T]\times \Omega \rightarrow \mathbb{R}^d$ be $\mathbb{F}$-adapted stochastic processes with continuous sample paths which satisfy that for all $t\in[0,T]$ it holds $\mathbb{P}$-a.s. that 
\begin{equation}
    \begin{aligned}
    Y_t = g(\xi+W_T) + \int^T_t f(Y_s, Z_s)\, ds - \int^T_t <Z_s, dW_s>_{\mathbb{R}^d}
    \end{aligned}
\end{equation}
Under suitable additional regularity assumptions on the nonlinearity $f$ we have that the nonlinear parabolic PDE is related to the BSDE in the sense that for all $t\in[0,T]$ it holds $\mathbb{P}$-a.s. that 
\begin{equation}
    Y_t = u(t,\xi+W_t)\in \mathbb{R} \quad \text{and} \quad Z_t = (\Delta_x u) (t,\xi+W_t)\in\mathbb{R}^d
\end{equation}
The first identity above is normally referred to as nonlinear Feynman-Kac Formula in the literature. 

\section{Formulation of examples}\label{appendix:pde}

\subsection{Hamilton Jacobi Bellman Equation}\label{appendix:HJB}

The equation takes the form of 
\begin{equation}
\begin{aligned}
\frac{\partial u}{\partial t}(t,x) + (\Delta_x u)(t,x) = \norm{(\nabla_x u)(t,x)}^2_{\mathbb{R}^d}
\end{aligned}
\end{equation}
The boundary condition of $g(x)=\log(\frac{1}{2}\left[1+\norm{x}^2_{\mathbb{R}^d}\right]$ such that $u(T,x)=g(x)$. We use $d=100$, $T=1$, $N=20$, $\mu(t,x)=0$ and $\sigma(t,x)=\sqrt{2}$ and solve for $u(0,\xi)$ where $\xi = (0,0,\ldots,0)\in \mathbb{R}^d$. In terms of hyperparameters, we choose batch size to be 512, learning rate to be $10^{-2}$. The exact solution of HJB is obtained via Monte Carlo simulations based on the formula \cite{chassagneux2016numerical}:
\begin{equation}
u(t,x) = -\frac{1}{\lambda} \log \left( \mathbb{E}\left[\exp \left(-\lambda g\left( x+\sqrt{2}W_{T-t}\right) \right)\right] \right)
\end{equation}
To determine the equilibrium of training, we use $\epsilon=5\times 10^{-4}$ and $P=200$. 

\subsection{Allen Cahn Equation}\label{appendix:AllenCahn}
\begin{equation}
\begin{aligned}
\frac{\partial u}{\partial t}(t,x) + u(t,x) -\left[u(t,x)\right]^3 + (\Delta_x u)(t,x)= 0
\end{aligned}
\end{equation}
The boundary condition is $g(x)=\left[2+\frac{2}{5}\norm{x}^2_{\mathbb{R}^d}\right]^{-1}$. We use $d=100$, $T=0.3$, $N=20$ and solve for $u(0,\xi)$ where $\xi = (0,0,\ldots,0)\in \mathbb{R}^d$. In terms of hyperparameters, we choose batch size to be 512, learning rate to be $5\times 10^{-4}$. The exact solution of Allen Cahn is not explicitly known. We approximate it by means of branching diffusion method \cite{henry2014numerical}. To determine the equilibrium of training, we use $\epsilon=100$ and $P=100$. 

\subsection{Black-Scholes Equation}\label{appendix:blackscholess}
\begin{equation}
\begin{aligned}
\frac{\partial u}{\partial t}(t,x) + f\left(t,x,u(t,x),\Bar{\sigma}{\rm diag}_{\mathbb{R}^{d\times d}}(x_1,\ldots,x_d)(\nabla_x u) (t,x)\right) + \Bar{\mu}\sum^d_{i=1}x_i \frac{\partial u}{\partial x_i}(t,x) + \frac{\Bar{\sigma}^2}{2}\sum^d_{i=1} \lVert x_i\rVert^2\frac{\partial ^2 u}{\partial x_i^2}(t,x)=0
\end{aligned}
\end{equation}
where $f(t,x,y,z)=\begin{aligned}  -R^l y-\frac{\Bar{\mu}-R^l}{\Bar{\sigma}}\sum^d_{i=1}z_i +   (R^b-R^l)
        \norm{\left[\frac{1}{\Bar{\sigma}}\sum^d_{i=1}z_i\right]-y}_\infty
    \end{aligned}$
    
The boundary condition is $g(x)={\rm max} \left\{ \left[ {\max\limits_{1\leq i \leq 100} } x_i \right]-120,0\right\} - 2 {\rm max } \left\{ \left[\max\limits_{1\leq i \leq 100} x_i \right] -150,0\right\}$. To determine the equilibrium of training, we use $\epsilon=10^{-2}$ and $P=100$.

\subsection{Burgers Type Equation}\label{appendix:burgers}
\begin{equation}
\begin{aligned}
\frac{\partial u}{\partial t}(t,x) + \frac{d^2}{2}(\Delta_x u)(t,x) + \left(u(t,x)-\frac{2+d}{2d}\right)\left(d\sum^d_{i=1}\frac{\partial u}{\partial x_i}(t,x)\right) = 0
\end{aligned}
\end{equation}
The boundary condition is
\begin{equation}
    g(x) = \frac{\exp (T+\frac{1}{d}\sum^d_{i=1} x_i}{(1+\exp(T+\frac{1}{d}\sum^d_{i=1}x_i))}
\end{equation}
We use $d=50$, $T=0.2$, $N=60$ and solve for $u(0,\xi)$ where $\xi = (0,0,\ldots,0)\in \mathbb{R}^d$. In terms of hyperparameters, we choose batch size to be 512, learning rate to be $10^{-2}$. The according exact solution can be found in Example 4.6 in Subsection 4.2 of \cite{chassagneux2014linear}. 
To determine the equilibrium of training, we use $\epsilon=10^{-3}$ and $P=300$. 

\subsection{Reaction Diffusion Equation}\label{appendix:reaction_diffusion}
\begin{equation}
\begin{aligned}
\frac{\partial u}{\partial t}(t,x) + \min\left\{ 1, \left[u(t,x)-\kappa-1-\sin(\lambda\sum^d_{i=1}x_i)\exp(\frac{\lambda^2d(t-T)}{2}) \right]^2\right\} + \frac{1}{2}(\Delta_x u)(t,x) = 0
\end{aligned}
\end{equation}
The boundary condition is
\begin{equation}
    g(x) = 1+\kappa+\sin(\lambda \sum^d_{i=1} x_i)
\end{equation}
We use $d=100$, $T=1$, $N=30$ and solve for $u(0,\xi)$ where $\xi = (0,0,\ldots,0)\in \mathbb{R}^d$. In terms of hyperparameters, we choose batch size to be 512, learning rate to be $10^{-2}$. The according exact solution can be found in in Subsection 6.1 of \cite{gobet2017adaptive}. 
To determine the equilibrium of training, we use $\epsilon=10^{-3}$ and $P=1200$. 

\subsection{Quadratically Growing Equation}\label{appendix:quadratically_growing}
\begin{equation}
\begin{aligned}
\frac{\partial u}{\partial t}(t,x) +\norm{(\nabla_x u)(t,x)}^2_{\mathbb{R}^d} + \frac{1}{2}(\Delta_x u)(t,x) = \frac{\partial \psi}{\partial t} (t,x) + \norm{(\nabla_x \psi)(t,x)}^2_{\mathbb{R}^d} + \frac{1}{2}(\Delta_x \psi)(t,x)
\end{aligned}
\end{equation}
where $\psi(t,x)=\sin\left(\left[ T-t+\norm{x}^2_{\mathbb{R}^d}\right]^\alpha\right)$ and the boundary condition is $g(x) = \sin\left(\norm{x}^{2\alpha}_{\mathbb{R}^d}\right)$.
We use $d=100$, $T=1$, $N=30$ and solve for $u(0,\xi)$ where $\xi = (0,0,\ldots,0)\in \mathbb{R}^d$. In terms of hyperparameters, we choose batch size to be 512, learning rate to be $10^{-2}$. The according exact solution can be found in in Section 5 of \cite{gobet2016linear}. To determine the equilibrium of training, we use $\epsilon=10^{-4}$ and $P=100$. 

\section{Evolution of loss and relative approximation error during training}
\label{appendix:training}

In this section we present some additional training plots for the numerical experiments. We compare the $L_1$ error, the loss level for each of the numerical experiments between our model and the DBSDE model.  

\begin{figure}[H]
    \centering 
\begin{subfigure}{0.25\textwidth}
  \includegraphics[width=\linewidth]{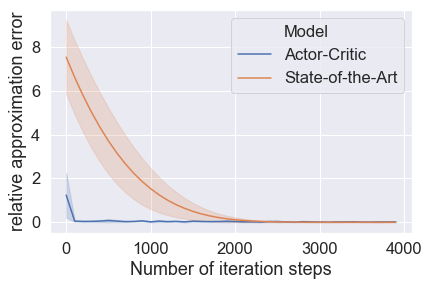}
  \caption{Allen Cahn}
  \label{fig:re_allen_cahn}
\end{subfigure}\hfil 
\begin{subfigure}{0.25\textwidth}
  \includegraphics[width=\linewidth]{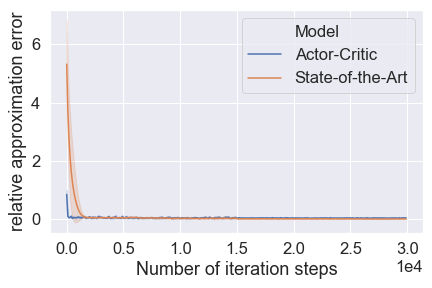}
  \caption{Burgers Type}
  \label{fig:re_burgers_type}
\end{subfigure}\hfil 
\begin{subfigure}{0.25\textwidth}
  \includegraphics[width=\linewidth]{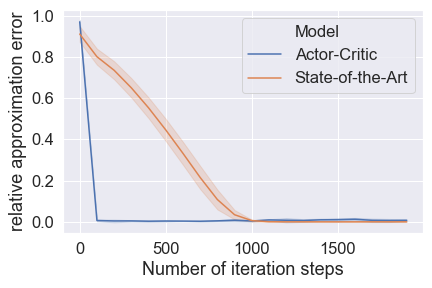}
  \caption{Hamilton Jacobian Bellman}
  \label{fig:re_HJB}
\end{subfigure}

\medskip
\begin{subfigure}{0.25\textwidth}
  \includegraphics[width=\linewidth]{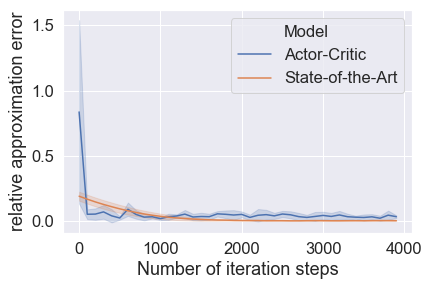}
  \caption{Pricing Option}
  \label{fig:re_pricing_option}
\end{subfigure}\hfil 
\begin{subfigure}{0.25\textwidth}
  \includegraphics[width=\linewidth]{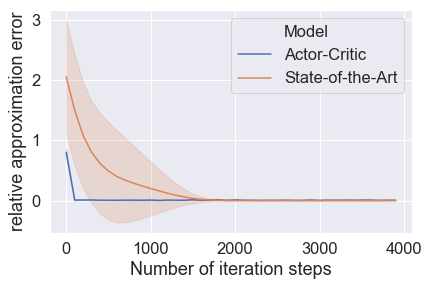}
  \caption{Quadratic Gradients}
  \label{fig:re_quadratic_gradients}
\end{subfigure}\hfil 
\begin{subfigure}{0.25\textwidth}
  \includegraphics[width=\linewidth]{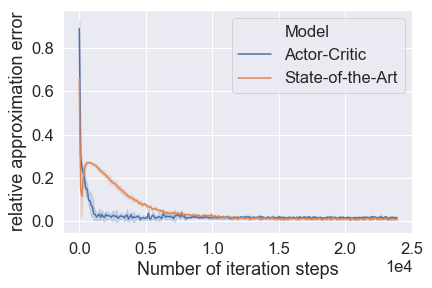}
  \caption{Reaction Diffusion}
  \label{fig:re_reaction_diffusion}
\end{subfigure}
\caption{Training history of relative approximation error}
\label{fig:re}
\end{figure}

\begin{figure}[H]
    \centering 
\begin{subfigure}{0.25\textwidth}
  \includegraphics[width=\linewidth]{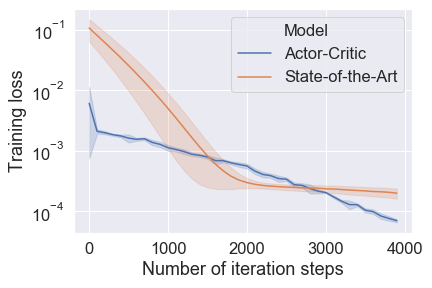}
  \caption{Allen Cahn}
  \label{fig:loss_allen_cahn}
\end{subfigure}\hfil 
\begin{subfigure}{0.25\textwidth}
  \includegraphics[width=\linewidth]{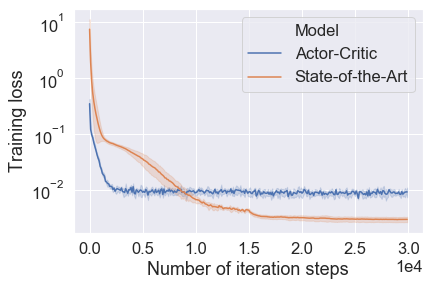}
  \caption{Burgers Type}
  \label{fig:loss_burgers}
\end{subfigure}\hfil 
\begin{subfigure}{0.25\textwidth}
  \includegraphics[width=\linewidth]{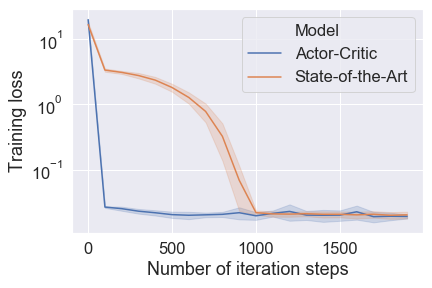}
  \caption{Hamilton Jacobian Bellman}
  \label{fig:loss_hjb}
\end{subfigure}

\medskip
\begin{subfigure}{0.25\textwidth}
  \includegraphics[width=\linewidth]{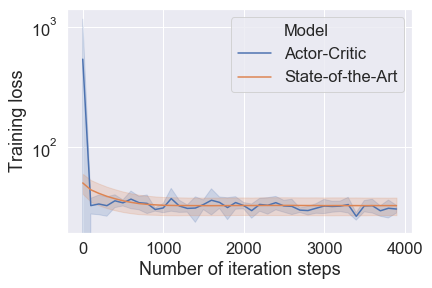}
  \caption{Pricing Option}
  \label{fig:loss_pricing}
\end{subfigure}\hfil 
\begin{subfigure}{0.25\textwidth}
  \includegraphics[width=\linewidth]{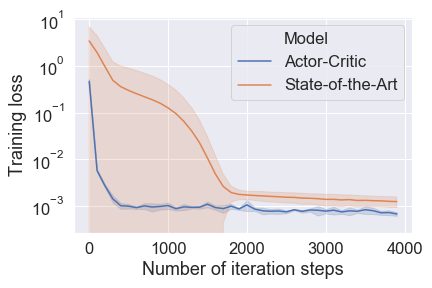}
  \caption{Quadratic Gradients}
  \label{fig:loss_quadratic_gradients}
\end{subfigure}\hfil 
\begin{subfigure}{0.25\textwidth}
  \includegraphics[width=\linewidth]{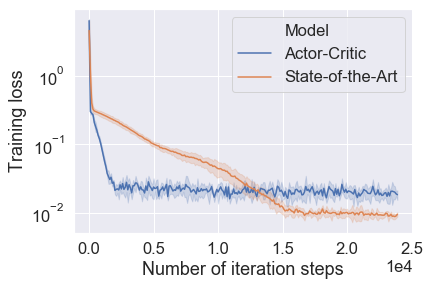}
  \caption{Reaction Diffusion}
  \label{fig:loss_reaction_diffusion}
\end{subfigure}
\caption{Training history of loss function}
\label{fig:loss_history}
\end{figure}
\end{document}